\documentclass{article}

\usepackage[final]{Vec2Instance_2020}

\usepackage{graphicx}
\graphicspath{ {./attachments/} }

\usepackage[utf8]{inputenc} 
\usepackage[T1]{fontenc}    
\usepackage{hyperref}       
\usepackage{url}            
\usepackage{booktabs}       
\usepackage{amsfonts}       
\usepackage{nicefrac}       
\usepackage{microtype}      

\title{Vec2Instance: Parameterization for Deep Instance Segmentation}

\author{%
  N. Lakmal Deshapriya\thanks{Corresponding author} \\
  Asian Institute of Technology, Thailand \\
  \texttt{lakmal@ait.ac.th} \\
  \And
  Matthew N. Dailey\\
  Asian Institute of Technology, Thailand \\
  \texttt{mdailey@ait.ac.th} \\
  \And
  Manzul Kumar Hazarika\\
  Asian Institute of Technology, Thailand \\
  \texttt{manzul@ait.ac.th} \\
  \And
  Hiroyuki Miyazaki\\
  Asian Institute of Technology, Thailand \\
  \texttt{miyazaki@ait.ac.th} \\
}

\begin{document}

\maketitle

\begin{abstract}
Current advances in deep learning is leading to human-level accuracy in computer vision tasks such as object classification, localization, semantic segmentation, and instance segmentation. In this paper, we describe a new deep convolutional neural network architecture called \mbox{Vec2Instance} for instance segmentation. \mbox{Vec2Instance} provides a framework for parametrization of instances, allowing convolutional neural networks to efficiently estimate the complex shapes of instances around their centroids. We demonstrate the feasibility of the proposed architecture with respect to instance segmentation tasks on satellite images, which have a wide range of applications. Moreover, we demonstrate the usefulness of the new method for extracting building foot-prints from satellite images. Total pixel-wise accuracy of our approach is 89\%, near the accuracy of the state-of-the-art Mask RCNN (91\%). \mbox{Vec2Instance} is an alternative approach to complex instance segmentation pipelines, offering simplicity and intuitiveness. The code developed under this study is available in the \mbox{Vec2Instance} GitHub repository, \url{https://github.com/lakmalnd/Vec2Instance}.
\end{abstract}

\section{Introduction}

Instance segmentation focuses on labeling each pixel of an image while maintaining instance awareness. It is one of the major open problems in computer vision, and it has many real-world applications. In this study, we have developed a new deep learning technique for instance segmentation. The feasibility of the approach is assessed using satellite images.

Many of the important applications of deep learning in computer vision, including instance segmentation and also 3D reconstruction, style transfer, and image generation, require generation of a parametrization of some aspects of an object in a high-dimensional space rather than identifying the object’s type. As an example, in the case of face generation tasks, a CNN should be able to generate images of faces rather than simply classifying the ``face'' object in the image as a face. We propose an approach to object generation by parametrizing discrete and continuous functions in higher dimensions (images, instance masks, 3D objects, etc.)  in a deep learning-friendly manner that we call \mbox{Vec2Instance}. We demonstrate the feasibility of \mbox{Vec2Instance} with an additional case study in another domain besides building instance masking: human face reconstruction by face parametrization.

\section{Background}

Current deep learning techniques such as YOLO (You Only Look Once) \citep{yolo} perform object localization tasks (bounding box estimation) very successfully. It seems intuitive that a similar deep neural network should be able perform object localization with complicated shapes rather than rectangular bounding boxes, solving the instance segmentation problem accurately and efficiently. In this section, we discuss previous approaches to CNN-based instance segmentation and the idea behind the YOLO object localization algorithm before explaining how we adapt it to the task of instance segmentation via instance parameterization.

\subsection{CNN-based Instance Segmentation}

To perform instance segmentation, a network has to perform three tasks, namely object localization, object classification, and masking of the object, while maintaining instance awareness. This makes it a challenging problem in computer vision. There have been several attempts to construct CNN-based architectures for instance segmentation, but Mask R-CNN \citep{maskRCNN} is one of most popular and successful attempts in this regards.

The key ideas of Mask R-CNN are based on the Faster R-CNN \citep{fasterRCNN} architecture, adding FCNs \citep{fcn} as an additional branch of the head network, enabling it to generate a mask for each region proposal. In Mask R-CNN, RPN and ROI pooling are used to generate cropped regions of interest that may contain objects. The Mask R-CNN network has two branches as follows. 

\begin{itemize}
\item Branch 1: bounding box regression with labeling (classification and localization)
\item Branch 2: masking of the object with a FCN (semantic segmentation)
\end{itemize}

Performing all three tasks (classification, localization, and semantic segmentation) in one network allows Mask R-CNN to perform instance segmentation in a successful manner. However, one disadvantage of Mask R-CNN is its complex nature.

CNN-based methods able to perform localization and classification without region proposals have also been proposed for the instance segmentation task. One of the problems they face is that CNNs use pooling operations to enable shift invariance, but that leads to a loss of positional information, especially near the head of network. This is particularly troublesome when it comes to instance segmentation, because positional information is critical for accurate instance segmentation.  Networks such as Deep Contour Awareness Networks (DCAN) \citep{dcan} make positional awareness explicit at the output of the CNN.

In the DCAN architecture, there are two branches. One branch of the network performs semantic segmentation of the scene, while the other performs semantic segmentation for the edges of objects. Combining edge segmentation and object segmentation in one network brings instance awareness to the segmentation task in the DCAN architecture. 

One additional related effort is the deep watershed transform for instance segmentation \citep{deep_watershed}. This approach goes beyond the DCAN approach by providing direct positional awareness in the head of the network rather than only edge awareness as in the DCAN architecture. Models with the deep watershed transform architecture attempt to learn two things: unit vectors pointing to / against the boundary, and the distance of each pixel from the edge.

\subsection{Object Localization: YOLO}

In the object localization task, a CNN is used to localize a particular object in an image by estimating the location and size of its bounding box. One of best-known approaches to CNN-based object localization is known as YOLO (You Only Look Once) \citep{yolo}. The most recent version of this algorithm is known as YOLO-V4 \citep{yolo_v4}, which is very accurate and efficient.

In YOLO, the object localization task is handled in an end-to-end manner. This allows a simpler streamlined architecture for the object localization task. As usual, in YOLO's backbone architecture, a CNN summarizes receptive fields with feature vectors. Those feature vectors are used to feed fully connected layers that regress parameters corresponding to the object localization task. Those parameters include

\begin{itemize}
\item Parameter 1 (objectness): the probability that an object is in the corresponding receptive field (0 if the receptive field is background and 1 if the receptive field contains an object).
\item Parameters 2--5 (bounding box parameters): the center, width, and height of the bounding box (bx, by, bw, bh).
\item Other parameters: the remaining parameters correspond to class labels. For example, if there are four object classes, there will be four parameters corresponding to the four classes.
\end{itemize}

To avoid multiple detections of the same object, YOLO uses non-maximum suppression, which refines a set of bounding boxes by removing bounding boxes overlapping other boxes with higher probabilities, leaving a set of non-overlapping bounding boxes with the highest probability.

\begin{figure}
\centering
\includegraphics[width=100mm]{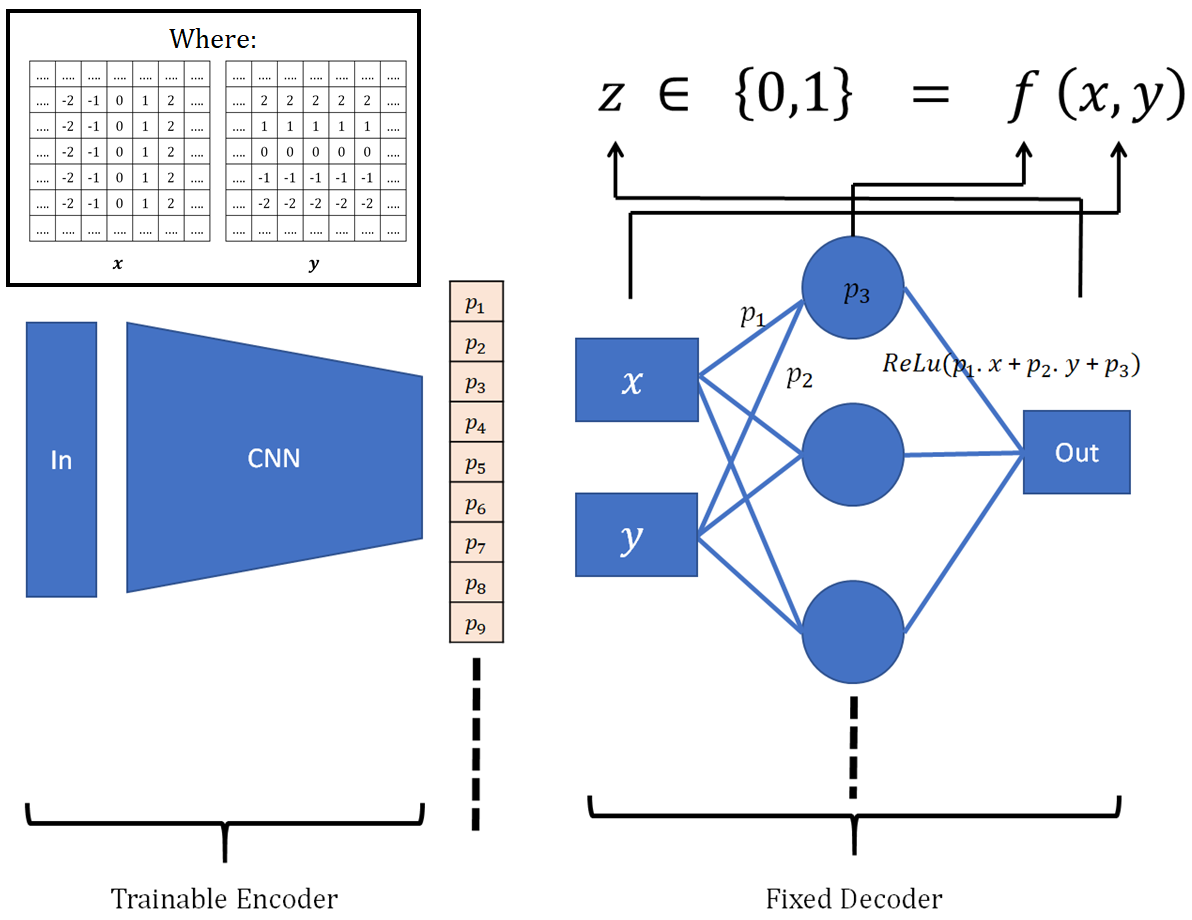} 
\caption{
\mbox{Vec2Instance} architecture.}
\label{fig:Fig_Vec2Instance}
\end{figure}

\section{Our Methodology}

In this section, the \mbox{Vec2Instance} architecture for instance segmentation is explained along with the intuition behind the method.

\subsection{\mbox{Vec2Instance} Intuition}

The key idea behind object localization in YOLO \citep{yolo} is to first estimate the centroid of each object and then to estimate the bounding box (center $x$, center $y$, width, and height) that encloses an instance. If we can specify a sufficiently rich differentiable function that similarly estimates the complex shape of an instance, going beyond the simple rectangular shape of the bounding box, we can use that function and the parameter regression idea to perform instance segmentation.

An instance mask is a multivariate function in two dimensions. The input to the multivariate function is a pixel location ($x$ and $y$ coordinate values), and the output of the multivariate function should be 1 if the location is inside the mask and 0 otherwise. The computational setup for modeling this family of functions has long been used by neural network practitioners in networks containing only fully connected layers. In the instance segmentation task, the input represents location information, and the output represents the instance mask. The weights and biases of a neural network modeling such a multivariate function relate location data to label values (the label value is one if the location is inside mask and zero otherwise). These weights and biases can be considered as parameters that represent an instance mask.

Now, once we have a way to parametrize an instance mask, a CNN can be used to learn those parameters. A fixed decoder without trainable parameters can be used to reconstruct an instance mask from the particular set of parameters output by the CNN for a specific centroid. Since the process of mask reconstruction from a parameter vector is based on fully differentiable units, we can use any gradient-based optimizer to train the entire encoder-decoder neural network architecture in an end-to-end manner with the criterion being the reconstruction error for the instance mask. A pictorial representation of this idea (\mbox{Vec2Instance}) is shown in Figure \ref{fig:Fig_Vec2Instance}. One way to think of the model is as a neural network (CNN) that predicts the weights and biases of another neural network (a vanilla multilayer perceptron).

The universal approximation theorem in neural networks \citep{universal_thm} states that feed-forward networks with a single hidden layer containing a sufficient number of neurons and sufficient training data can model any function to a given level of accuracy (in our scenario, any 2D mask function). An instance segmentation application based on the universal approximation theorem implies that we can theoretically approximate any instance shape as long as we use a sufficiently-large hidden layer.

Based on the above intuition, we propose a new neural network architecture for the instance segmentation task comprising two CNNs. The first CNN estimates centroids of instances, and the second CNN preforms instance segmentation around each centroid. One special property of both CNNs is the use of the dilated convolution operations \citep{dilated_conv}. Dilated convolutions allow an increase in receptive field size without losing spatial resolution. This will be advantageous for instance segmentation due to the small image tiles we are using.

\subsection{Centroid Estimation CNN}

For centroid estimation, we propose a simple conventional CNN with convolutional and max-pooling layers. Sample input and output images for the centroid estimation CNN are shown in first and second rows of Figure \ref{fig:Fig_Results_centroid_CNN}. The input images are RGB satellite image tiles, and the output images are binary images (if a pixel is a centroid of a building, its target is 1; otherwise, its target is 0). A list of layers in the proposed centroid estimation CNN is given below.

\begin{itemize}
\item 32 3$\times$3 convolutions with ReLu activations
\item Dropout with 0.25 rate
\item 32 3$\times$3 convolutions with ReLu activations
\item 2$\times$2 Max pooling

\item 32 3$\times$3 convolutions with ReLu activations using 2x2 dilation
\item Dropout with 0.25 rate
\item 32 3$\times$3 convolutions with ReLu activations using 2x2 dilation
\item 2$\times$2 Max pooling

\item 32 3$\times$3 convolutions with ReLu activations using 2x2 dilation
\item Dropout with 0.25 rate
\item 32 3$\times$3 convolutions with ReLu activations using 2x2 dilation
\item 2$\times$2 Max pooling

\item 32 3$\times$3 convolutions with ReLu activations using 2x2 dilation
\item Dropout with 0.25 rate
\item 32 3$\times$3 convolutions with ReLu activations using 2x2 dilation

\item 32 1$\times$1 convolutions with ReLu activations
\item 32 1$\times$1 convolutions with ReLu activations
\item 1 1$\times$1 convolutions with ReLu activations
\end{itemize}

\begin{figure}
\centering
\includegraphics[width=120mm]{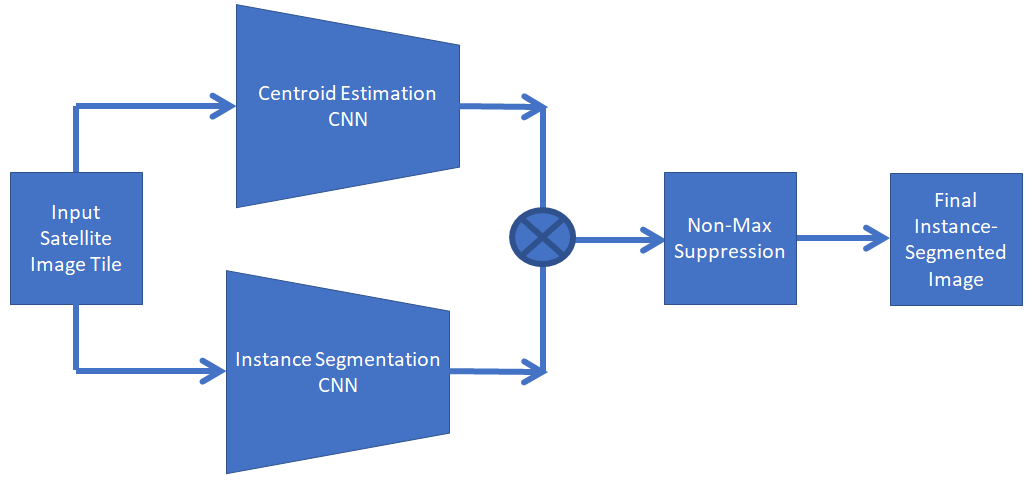} 
\caption{
Workflow of prediction stage.}
\label{fig:Fig_PredictWorkflow}
\end{figure}

\subsection{Instance Segmentation CNN}

In the case of the instance segmentation CNN, each individual object patch and its segmentation is provided to the network in the training stage. An RGB image centered on a single instance is provided as input, while the segmented instance of the corresponding object is provided as the output image. The mask generation decoder consists of a fully-connected neural network with a single 64-unit hidden layer. This decoder consists of 257 weights and biases in total (2$\times$64 weights and 64 biases in the fully connected hidden layer and 64$\times$1 weights and 1 bias in the last fully connected layer). The 257 weights and biases of the decoder, which parametrize building masks, are learned through a CNN.

Since the network’s convolution operations are limited to actual object instances, the performance of the network should be much higher than performing convolutions over a large image tile containing large regions of background (areas without buildings). This also helps to overcome issues with overlapping instances. Sample input and output images for the instance segmentation CNN are shown in the first and second rows of Figure \ref{fig:Fig_Results_instance_CNN}.

A list of layers in the instance segmentation CNN is shown below.

\begin{itemize}
\item 32 3$\times$3 convolutions with ReLu activations
\item Dropout with 0.25 rate
\item 32 3$\times$3 convolutions with ReLu activations
\item 2$\times$2 Max pooling

\item 32 3$\times$3 convolutions with ReLu activations using 2x2 dilation
\item Dropout with 0.25 rate
\item 32 3$\times$3 convolutions with ReLu activations using 2x2 dilation
\item 2$\times$2 Max pooling

\item 64 3$\times$3 convolutions with ReLu activations using 2x2 dilation
\item Dropout with 0.25 rate
\item 64 3$\times$3 convolutions with ReLu activations using 2x2 dilation
\item 2$\times$2 Max pooling

\item 64 3$\times$3 convolutions with ReLu activations using 2x2 dilation
\item Dropout with 0.25 rate
\item 64 3$\times$3 convolutions with ReLu activations using 2x2 dilation

\item 64 1$\times$1 convolutions with ReLu activations
\item 64 1$\times$1 convolutions with ReLu activations
\item 257 1$\times$1 convolutions with Linear activations (257 element vector in the center is extracted as parameter vector)
\item Decoder: a single hidden layer containing 64 hidden units. Input is an $x,$ $y$ coordinate pair; the output is a single binary variable. The total of 257 weights and biases are learned by the CNN part of the network.
\end{itemize}

\begin{figure}
\centering
\includegraphics[width=120mm]{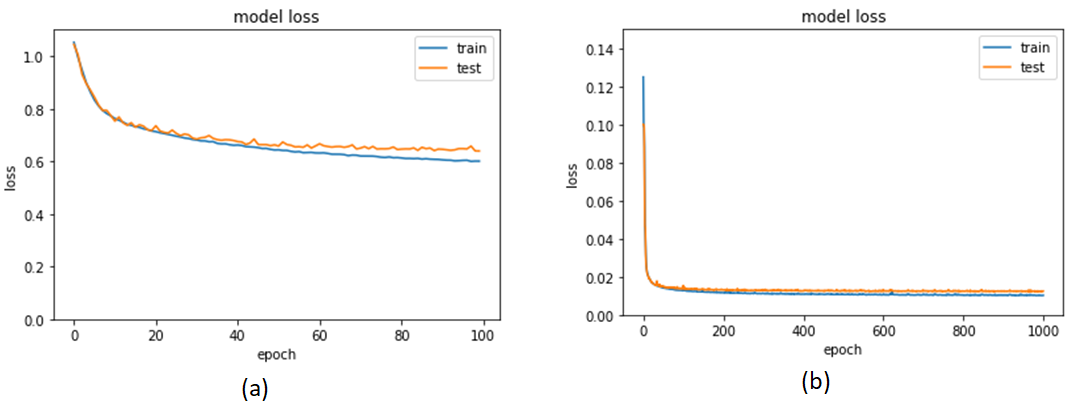}
\caption{
Loss curves for (a) centroid estimation CNN and (b) instance segmentation CNN.}
\label{fig:Fig_LossCurve}
\end{figure}

\subsection{Training and Prediction}

In the training stage, we train the two CNNs separately. The CNN that estimates centroids is trained with tiles from satellite images. The CNN that conducts instance segmentation is trained with actual instances of individual buildings. In both CNNs, the root mean squared error (RMSE) between the reconstructed centroid or instance and the ground truth is used for the loss function. Both networks are trained with the Adam optimizer \citep{adam}.

In the prediction stage, a tile of a satellite image is provided to both the centroid estimation CNN and the instance segmentation CNN. Instances are generated for the predicted centroid points, ignoring any point not predicted as centroid points. Overlapping segments are removed using the usual non maximum suppression routine, leaving instance masks with the highest probability among the original possibly overlapping instance masks. The overall workflow of the prediction stage is shown in Figure \ref{fig:Fig_PredictWorkflow}.

\section{Experimental Evaluation}

This section demonstrates our implementation of \mbox{Vec2Instance} executed against the SpaceNet challenge AOI 2 (Vegas) building footprint dataset \citep{spacenet},  which mainly consists of satellite images and building footprint data as the ground truth data.

\subsection{Pre-processing}

First, the following pre-processing steps were conducted to generate a machine learning readable dataset from the AOI 2 (Vegas) dataset.

\begin{itemize}
\item Convert GeoJSON building polygons to raster data corresponding to satellite image tiles (650$\times$650 pixel image tiles).
\item Resample all images to 256$\times$256 pixels.
\item Remove partially captured images. We also removed partially-captured buildings from the dataset for instance segmentation, because a partially-captured building may not include enough information estimate the complete shape of the building. However, we retain the partially-captured shapes for the centroid estimation CNN.
\item Remove image tiles without a significant number of buildings
\item Remove image tiles with buildings whose dimensions are larger than 64$\times$64 pixels (the designed image input and output size for the instance segmentation CNN).
\end{itemize}

\begin{figure}
\centering
\includegraphics[width=120mm]{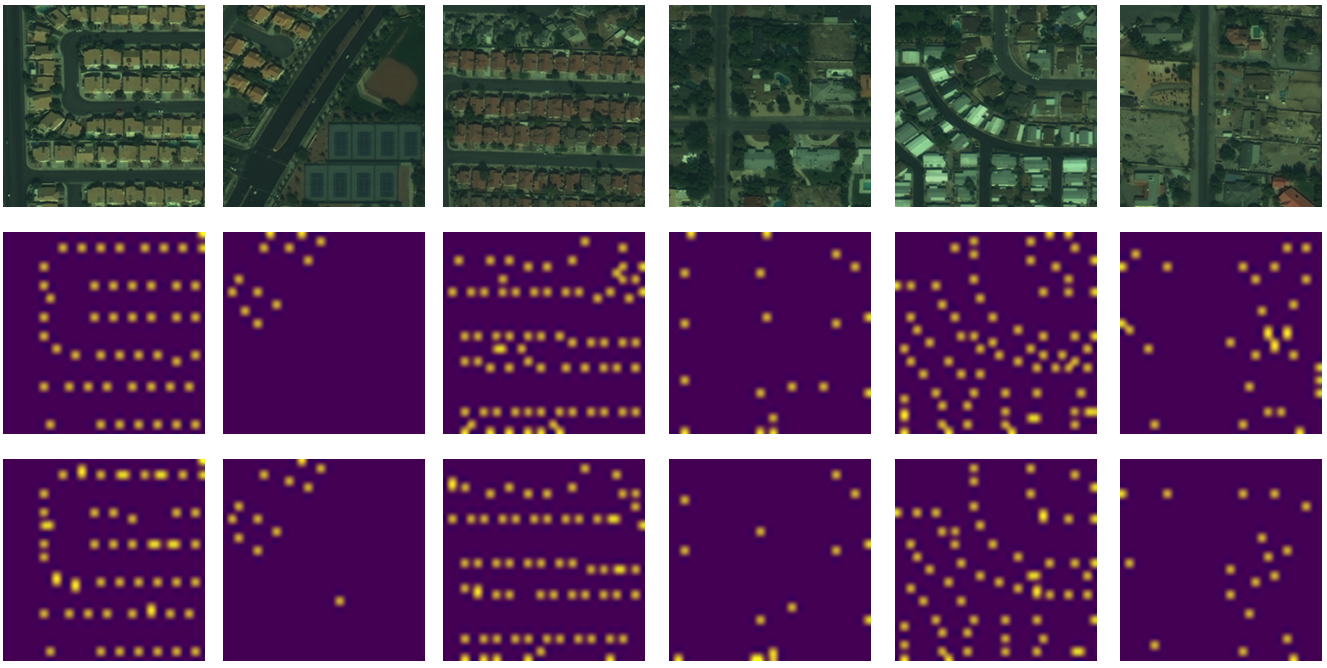}
\caption{
Sample experimental centroid estimation results for the test dataset. Input images are in first row, ground truth images are in second row, and results of centroid estimation are in third row.}
\label{fig:Fig_Results_centroid_CNN}
\end{figure}

\begin{figure}
\centering
\includegraphics[width=120mm]{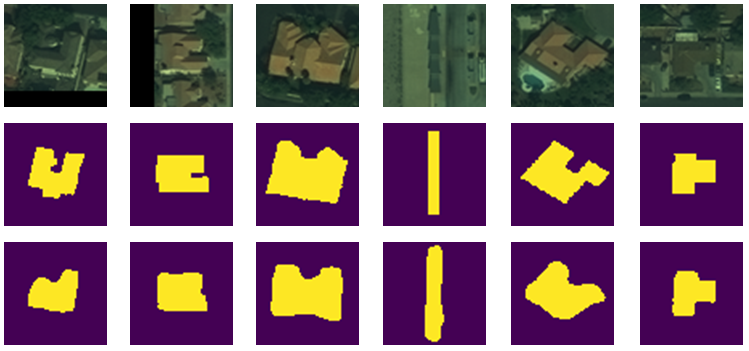}
\caption{
Sample experimental instance segmentation results for the test dataset. Input images are in first row, ground truth images are in second row, and results of instance segmentation are in third row.}
\label{fig:Fig_Results_instance_CNN}
\end{figure}

\setlength{\tabcolsep}{4pt}
\begin{table}
\begin{center}
\caption{
Confusion matrix for centroid estimation CNN over test dataset after thresholding CNN output then comparing predicted and ground truth centroids with distance threshold of 0 in the downscaled (64$\times$64) output image. Per-pixel total accuracy over all test centroids was 98\%.
}
\label{table:confusion_matrix_centroid_cnn}
\begin{tabular}{lll}
\hline\noalign{\smallskip}
 $\qquad\qquad$& Non-building pixels & Building pixels \\
 $\qquad\qquad$& (Predicted) & (Predicted)\\
\noalign{\smallskip}
\hline
\noalign{\smallskip}
	  Non-building pixels (Actual) & 95\% & 1\% \\
      Building pixels (Actual) & 1\% & 3\% \\
\hline
\end{tabular}
\end{center}
\end{table}
\setlength{\tabcolsep}{1.4pt}

\setlength{\tabcolsep}{4pt}
\begin{table}
\begin{center}
\caption{
Confusion matrix for instance segmentation CNN over test dataset after thresholding CNN output. Per-pixel total accuracy over all test buildings was 98\%.
}
\label{table:confusion_matrix_instance_cnn}
\begin{tabular}{lll}
\hline\noalign{\smallskip}
 $\qquad\qquad$& Non-building pixels & Building pixels \\
 $\qquad\qquad$& (Predicted) & (Predicted)\\
\noalign{\smallskip}
\hline
\noalign{\smallskip}
	  Non-building pixels (Actual) & 89\% & 1\% \\
      Building pixels (Actual) & 1\% & 9\% \\
\hline
\end{tabular}
\end{center}
\end{table}
\setlength{\tabcolsep}{1.4pt}

\setlength{\tabcolsep}{4pt}
\begin{table}
\begin{center}
\caption{
Confusion matrix for overall approach over test dataset. Per-pixel total accuracy over all test tiles was 89\%.
}
\label{table:confusion_matrix_overall}
\begin{tabular}{lll}
\hline\noalign{\smallskip}
 $\qquad\qquad$& Non-building pixels & Building pixels \\
 $\qquad\qquad$& (Predicted) & (Predicted)\\
\noalign{\smallskip}
\hline
\noalign{\smallskip}
	  Non-building pixels (Actual) & 72.1\% & 5.2\% \\
      Building pixels (Actual) & 5.7\% & 17\% \\
\hline
\end{tabular}
\end{center}
\end{table}
\setlength{\tabcolsep}{1.4pt}

\subsection{Training Details}

Both networks were trained on a single GPU (GEFORCE GTX 1080 Ti). We use the Keras deep learning framework with the TensorFlow backend for the implementation. Training details for the centroid estimation CNN are as below, and the loss profile for the training and test sets are shown in Figure \ref{fig:Fig_LossCurve}(a).

\begin{itemize}
\item Number of epochs: 100 
\item Total training time: 10 minutes
\item Loss function: weighted root mean squared error (0.66 weight on centroids and 0.33 weight on non-centroids)
\item Optimizer: Adam \citep{adam}
\item Batch size: 50
\item Input image size: 256$\times$256 RGB
\item Output image size: 32$\times$32 images with centroids
\item Total images: 1,884
	\begin{itemize}
	\item Training set: 2/3 of dataset (1,236 images)
	\item Test set: 1/3 of dataset (648 images)
	\end{itemize}
\end{itemize}

Training details for the instance segmentation CNN are as below, and the loss profile for the training and test sets are shown in Figure \ref{fig:Fig_LossCurve}(b).

\begin{itemize}
\item Number of epochs: 1000
\item Total training time: 4 hours and 10 minutes
\item Loss function: root mean squared error
\item Optimizer: Adam \citep{adam}
\item Batch size: 500
\item Input image size: 64$\times$64 RGB
\item Output image size: 64$\times$64 images with building instance masks
\item Total images: 53,594
	\begin{itemize}
	\item Training set: all buildings in 2/3 of image tiles (35,075 images)
	\item Test set: all buildings in 1/3 of image tiles (18,519 images)
	\end{itemize}
\end{itemize}

\begin{figure}
\centering
\includegraphics[width=120mm]{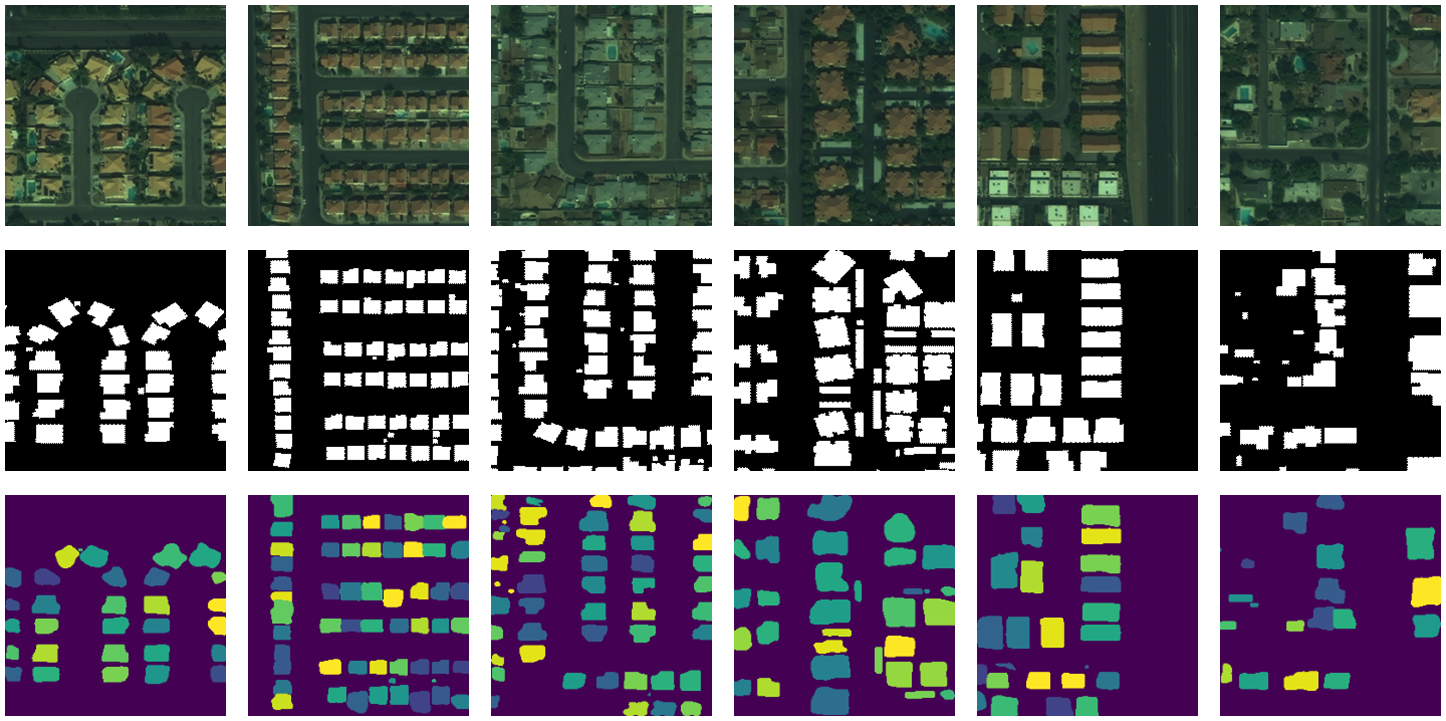}
\caption{
Sample experimental results for the test dataset. Input images are in first row, ground truth images are in second row, and combined results after non maximum suppression are in third row.}
\label{fig:Fig_Results_Good_Bad}
\end{figure}

\begin{figure}[t]
\centering
\includegraphics[width=120mm]{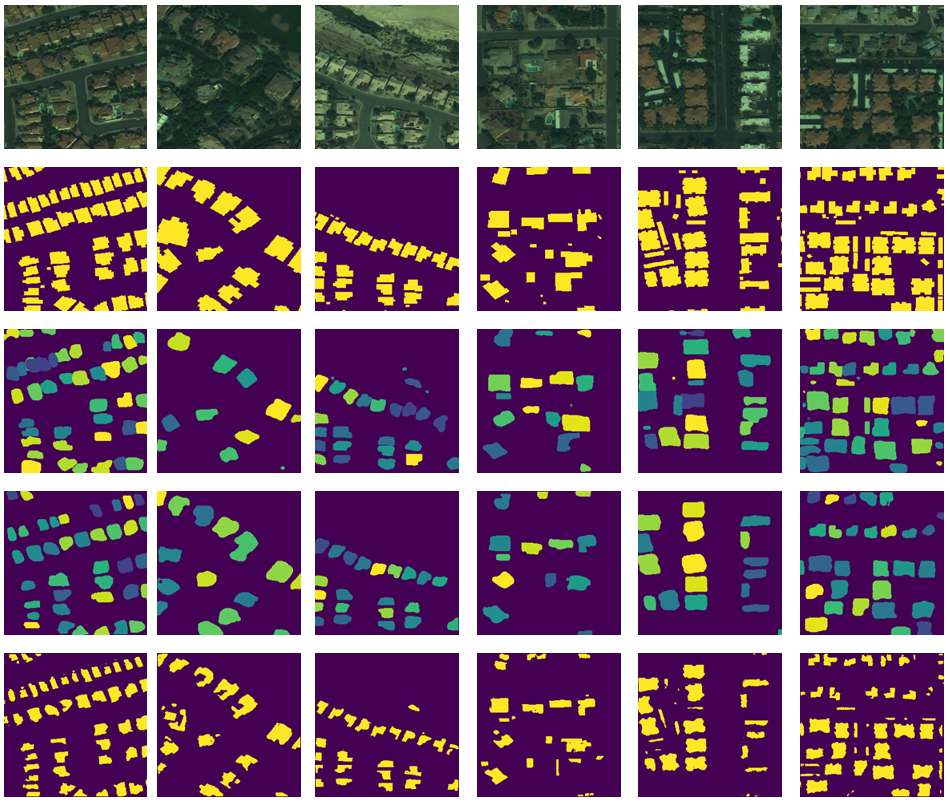}
\caption{
Comparison of results. Input images are in first row, ground truth images are in second row, \mbox{Vec2Instance} results are in third row, Mask RCNN results are in fourth row, and U-NET results are in fifth row.}
\label{fig:Fig_Results_comparison}
\end{figure}

\setlength{\tabcolsep}{4pt}
\begin{table}
\begin{center}
\caption{
Accuracy comparison on test set.
}
\label{table:comparison_accuracy}
\begin{tabular}{llll}
\hline\noalign{\smallskip}
Method $\qquad\qquad$& Overall Pixel-wise & Intersection over & Training Time\\
		$\qquad\qquad$& Accuracy & Union (IoU) & (Hours)\\
\noalign{\smallskip}
\hline
\noalign{\smallskip}
      \mbox{Vec2Instance} & 89\% & 61\% & 4.3  \\
      Mask R-CNN & 91\% & 65\% & 8.6 \\
      U-Net & 96\% & 84\% & 2.5 \\
\hline
\end{tabular}
\end{center}
\end{table}
\setlength{\tabcolsep}{1.4pt}

\section{Experimental Results}

Figures \ref{fig:Fig_Results_centroid_CNN} and \ref{fig:Fig_Results_instance_CNN} show sample results on the test dataset for the centroid estimation CNN and the instance segmentation CNN, respectively. Results after combining the centroid estimation CNN and the instance segmentation CNN with non maximum suppression are shown in Figures \ref{fig:Fig_Results_Good_Bad}. Tables \ref{table:confusion_matrix_centroid_cnn} and \ref{table:confusion_matrix_instance_cnn} present confusion matrices for assessment of the accuracy of the centroid estimation CNN and the instance segmentation CNN. Table \ref{table:confusion_matrix_overall} presents a confusion matrix for assessment of the accuracy of the overall system. The centroid estimation CNN and the instance segmentation CNN performed very well separately, both achieving a 98\% total pixel-wise accuracy. On the other hand, the total pixel-wise accuracy of the combined system was 89\%, quite a bit lower than the performance of the individual networks.

The loss of accuracy in the combined system is mainly due to missing buildings in the results compared to the ground truth (false negatives), and not due to detecting buildings in non-building locations (false positives). Some of the missing buildings are those with odd shapes, not the predominant building type in the study area. Others include those partially covered with trees and those without strong edges distinguishing them from the surroundings. In these locations, our model performance was weaker than in places with well-organized settlements.

We compared \mbox{Vec2Instance} with Mask RCNN \citep{maskRCNN}, which is a current state-of-the-art method in instance segmentation, as well as U-Net \citep{unet}, which is a current state-of-the-art method for semantic segmentation. For Mask R-CNN, we used the MatterPort Mask R-CNN implementation \citep{matterport_maskrcnn} based on Python 3, Keras, and TensorFlow. ResNet-50 \citep{resnet} was used as a backbone, and parameters were initialized to pre-trained values from the COCO dataset \citep{coco-dataset}. Only stage 3 and the upper stages of ResNet-50 were fine-tuned, keeping other parameters fixed. In terms of the number of parameters, the Mask R-CNN is very large compared to \mbox{Vec2Instance}.  We designed the U-Net network to have 421,793 parameters, nearly the same number of parameters as in \mbox{Vec2Instance} (166,305 parameters in the centroid estimation CNN and 182,945 parameters in the instance segmentation CNN).

Figure \ref{fig:Fig_Results_comparison} shows sample results for the test dataset from the three different approaches for comparison. Table \ref{table:comparison_accuracy} presents an accuracy assessment, including overall pixel-wise accuracy and intersection over union (IoU) for our approach, Mask R-CNN, and U-Net. While \mbox{Vec2Instance} achieved accuracy close to that of Mask R-CNN, the U-Net semantic segmentation network outperformed both \mbox{Vec2Instance} and Mask R-CNN.

\section{Conclusion and Discussion}

Through this research, we have introduced an alternative approach to instance segmentation by providing a framework for parametrization of instance masks (\mbox{Vec2Instance}) in a neural network-friendly manner. Based on the results of this case study, \mbox{Vec2Instance} parametrization reaches accuracy close to state-of-the-art instance segmentation methods such as Mask RCNN. The Mask RCNN implementation used for comparison here is a very large network with 44 million parameters compared to the 0.35 million total parameters in our approach. This gives Mask RCNN an advantage, but the complicated nature of Mask RCNN compared to \mbox{Vec2Instance} is also a disadvantage when it comes to implementation. On the other hand, the U-Net network used here has approximately the same number of parameters as in our approach, but it achieves nearly perfect results, performing much better than either our approach or Mask RCNN. However, not having in-built instance awareness is a disadvantage of the U-Net semantic segmentation network. Post processing is required with U-Net to give it instance awareness.

\begin{figure}
\centering
\includegraphics[width=80mm]{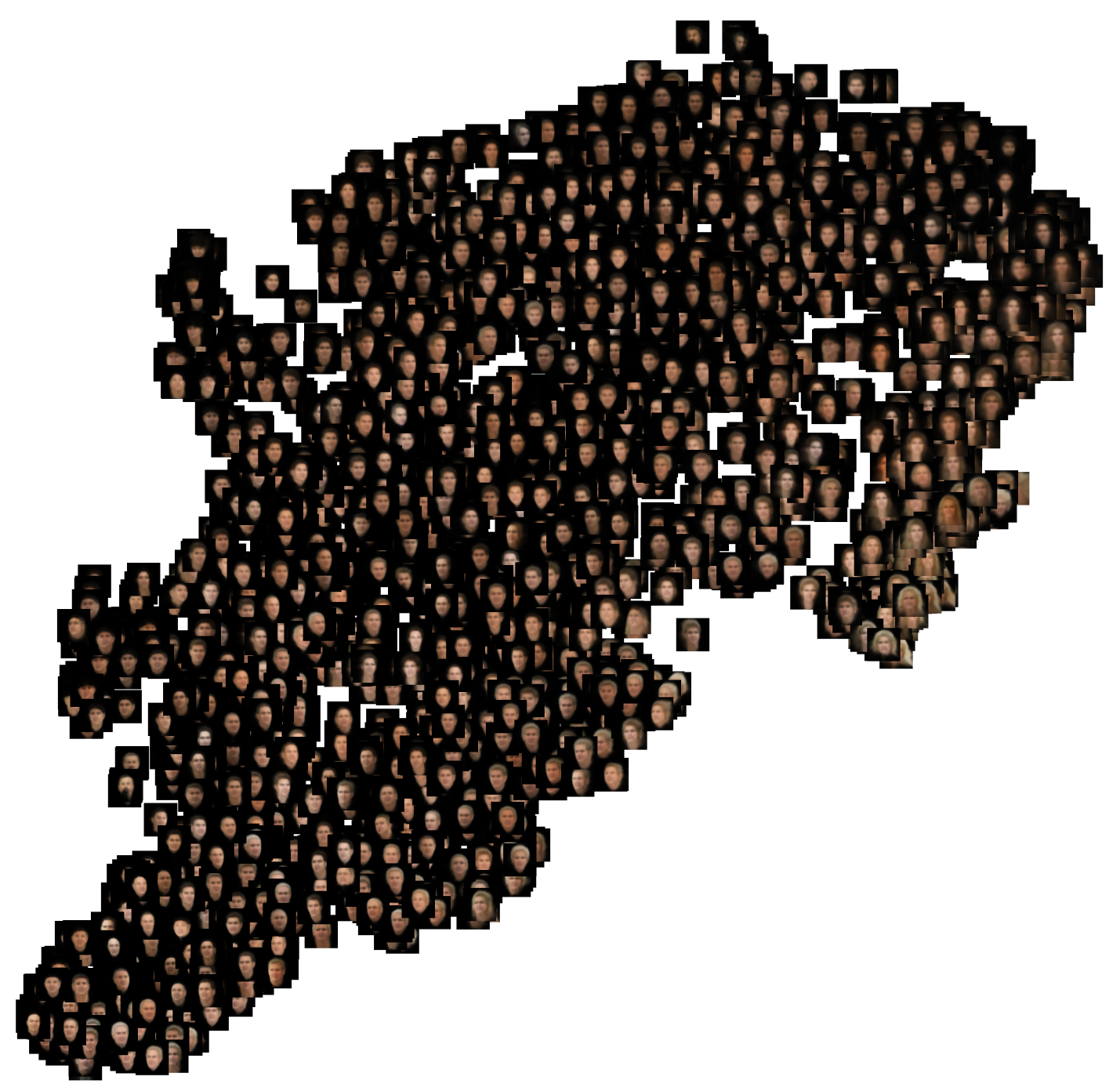}
\caption{
Visualization of \mbox{Vec2Instance} parameter space for 2D faces after applying dimensionality reduction.}
\label{fig:Fig_tSNE_Faces}
\end{figure}

\begin{figure}
\centering
\includegraphics[width=70mm]{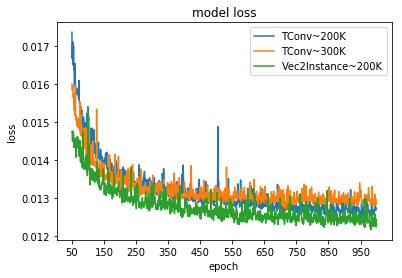}
\caption{
Loss curves for the test dataset for a decoder using of parametrization \mbox{Vec2Instance} and two decoders with transpose convolutions.}
\label{fig:Fig_LossCurve_TConv_Our}
\end{figure}

\mbox{Vec2Instance} can be used to parametrize a variety of objects, even entire images (human faces). Human faces can be considered as multivariate functions in two dimensions. The input to a function in this family is a location ($x$ and $y$ coordinates), and the output can be considered a tuple of pixel intensity values (RGB intensity values). With this intuition, as in our instance segmentation approach, a CNN can be used to estimate parameters that represent a face, and then a fixed decoder (without any trainable parameters) can be used to generate the face image from the estimated parameters. The entire network can be trained end-to-end to reconstruct input images at the end of the network. Experimental results on the ``Part Labels Database'' dataset \citep{face_dataset}, a subset of the “Labeled Faces in the Wild” (LFW) dataset consisting of 2,927 masked face images, are shown in Figure \ref{fig:Fig_tSNE_Faces}. The parameter space of faces (represented by 2,307-element vectors) learned through CNN is visualized in Figure \ref{fig:Fig_tSNE_Faces} in 2D after applying T-SNE dimensionality reduction to two dimension \citep{t_sne}. The faces are parametrized by a vanilla multilayer perceptron, comprising 2,307 weights and biases.

Similarly, \mbox{Vec2Instance} can be extended to more than two dimensions. For example, 3D shape reconstruction from a single RGB image can be performed with our parametrization approach as well. Some existing work includes the idea of the implicit decoder for 3D reconstruction \citep{shapenet}. Even though shapes are not parametrized in the same way as in our approach, providing coordinates to a decoder implicitly improves the 3D reconstruction. Theoretically, the idea of parametrization can be used to parametrize any object (discrete or continuous) in any number of dimensions.

Finally, \mbox{Vec2Instance} allows us to eliminate transpose convolution operations \citep{fcn} consisting of trainable parameters common in segmentation networks, reducing the total number of trainable parameters in the network. Since all instances are generated through a centroid vector, the centroid vector then becomes a bottleneck. With this bottleneck, the \mbox{Vec2Instance} decoder performs better than a transpose convolution decoder. This is demonstrated in Figure \ref{fig:Fig_LossCurve_TConv_Our} by loss curves over the test dataset of a decoder with \mbox{Vec2Instance} (200,000 total trainable parameters) and two decoders with transpose convolutions (200,000 and 300,000 total trainable parameters).

\bibliographystyle{unsrtnat}
\bibliography{references}

\end{document}